\documentclass[letterpaper, 10pt, conference]{ieeeconf}      

\IEEEoverridecommandlockouts                              

\usepackage{etex} 
\usepackage{tikz}
\usetikzlibrary{decorations.markings}
\usetikzlibrary{%
    decorations.pathreplacing,%
    decorations.pathmorphing%
}
\usepackage{gensymb}

\usepackage{caption}
\usepackage{array}
\newcolumntype{P}[1]{>{\centering\arraybackslash}p{#1}}
\newcolumntype{M}[1]{>{\centering\arraybackslash}m{#1}}

\usepackage{mathtools}

\usepackage{array}
\newcolumntype{P}[1]{>{\centering\arraybackslash}p{#1}}
\newcolumntype{M}[1]{>{\centering\arraybackslash}m{#1}}

\usepackage{lipsum}
\usepackage[utf8]{inputenc}
\usepackage{makeidx}
\usepackage{amsmath,bm}
\usepackage{siunitx}
\usepackage{amssymb}
\usepackage{color, colortbl}
\usepackage{ifpdf}
\usepackage{wrapfig}
\usepackage{graphicx}
\usepackage{sidecap}
\usepackage[]{threeparttable}
\usepackage{multirow}
\usepackage{multicol}
\usepackage{relsize}
\usepackage{algorithm}
\usepackage[noend]{algpseudocode}
\usepackage{pgfplots}
\pgfplotsset{compat=1.18}
\usepackage{mathrsfs}
\usepackage{bbm}
\usepackage{mathtools}
\usepackage{upgreek}
\usepackage{trfsigns}
\usepackage{arcs}
\usepackage{pgf}
\usepackage{tikz}
\usepackage{lscape}
\usepackage{array, makecell} %
\usepackage{todonotes}
\usepackage{caption}
\usepackage{subcaption}
\usepackage{svg}
\usepackage{cite}
\usepackage{soul}
\usepackage{comment}
\usepackage{import} 
\usepackage{amsmath}
\usepackage{amssymb}

\usepackage[section]{placeins} 
\usepackage{float}             
\usepackage{balance}
\usepackage{stfloats}
\usepackage{fancyhdr}

\usetikzlibrary{shapes.geometric,arrows,automata,positioning}
\usepackage[figuresright]{rotating}
\usepackage[stable, hang]{footmisc}
\usepackage{xparse,mathtools}
\usepackage{xstring}
\ifpdf
  \DeclareGraphicsExtensions{.pdf,.jpg,.png,.ps,.eps}
\else
  \DeclareGraphicsExtensions{.eps,.ps,.png,.jpg,.pdf}
\fi
\usepackage[normalem]{ulem}
\DeclareMathAlphabet{\mathbfsf}{\encodingdefault}{\sfdefault}{bx}{sl}
\DeclareMathAlphabet{\mathitsf}{\encodingdefault}{\sfdefault}{m}{sl}

\ifx\isieee 

	\renewcommand{\eqref}[1]{(\ref{#1})}

\else

\fi

\tikzstyle{startstop} = [rectangle, rounded corners, minimum width=3cm, minimum height=1cm,text centered, draw=black, thick, fill=red!30]
\tikzstyle{io} = [trapezium, trapezium left angle=70, trapezium right angle=110, minimum width=3cm, minimum height=1cm, text centered, draw=black, thick, fill=blue!30]
\tikzstyle{process} = [rectangle, minimum width=3cm, minimum height=1cm, text centered, draw=black, thick, fill=orange!30]
\tikzstyle{decision} = [diamond, aspect=3, minimum width=3cm, minimum height=1cm, text centered, draw=black, thick, fill=green!30]
\tikzstyle{arrow} = [thick,->,>=stealth]
\tikzstyle{noarrow} = [thick,-]
\tikzstyle{doublearrow} = [thick, double distance=1pt, ->, >=stealth]
\tikzstyle{doublenoarrow} = [thick, double distance=1pt]
\tikzstyle{innerwhite} = [thin, white, line width=1pt]
\tikzset{state/.style={
           rectangle,
           rounded corners,
           draw=black, very thick,
           minimum height=2em,
           inner sep=2pt,
           },}
\tikzstyle{start} = [circle, inner sep=0.06cm, draw=black, fill=black]      

\tikzstyle{gain} = [draw, very thick, fill=white, regular polygon, regular polygon sides=3, shape border rotate=30, minimum height=3em, minimum width=3em]
\tikzstyle{bigblock} = [draw, thick, fill=white, rectangle, 
minimum height=4em, minimum width=6em]
\tikzstyle{smallblock} = [draw, thick, fill=white, rectangle, 
minimum height=1em, minimum width=1em]
\tikzstyle{medianblock} = [draw, thick, fill=white, rectangle, 
minimum height=3em, minimum width=3em]
\tikzstyle{bigfcnblock} = [draw, thick, fill=white, rectangle, 
minimum height=4em, minimum width=6em, double, double distance=0.5mm]
\tikzstyle{smallfcnblock} = [draw, thick, fill=white, rectangle, 
minimum height=1em, minimum width=1em, double, double distance=0.5mm]
\tikzstyle{medianfcnblock} = [draw, thick, fill=white, rectangle, 
minimum height=3em, minimum width=3em, double, double distance=0.5mm]
\tikzstyle{invisiblenode} = [draw=none, minimum size=0pt]
\tikzstyle{sum} = [draw, thick, fill=white, circle, node distance=1cm]
\makeatletter
\pgfdeclareshape{record}{
	\inheritsavedanchors[from={rectangle}]
	\inheritbackgroundpath[from={rectangle}]
	\inheritanchorborder[from={rectangle}]
	\foreach \x in {center,north east,north west,north,south,south east,south west}{
		\inheritanchor[from={rectangle}]{\x}
	}
	\foregroundpath{
		\pgfpointdiff{\northeast}{\southwest}
		\pgf@xa=\pgf@x \pgf@ya=\pgf@y
		\northeast
		\pgfpathmoveto{\pgfpoint{0}{0.33\pgf@ya}}
		\pgfpathlineto{\pgfpoint{0}{-0.33\pgf@ya}}
		\pgfpathmoveto{\pgfpoint{0.33\pgf@xa}{0}}
		\pgfpathlineto{\pgfpoint{-0.33\pgf@xa}{0}}
		\pgfpathmoveto{\pgfpointadd{\southwest}{\pgfpoint{-0.33\pgf@xa}{-0.6\pgf@ya}}}
		\pgfpathlineto{\pgfpointadd{\southwest}{\pgfpoint{-0.5\pgf@xa}{-0.6\pgf@ya}}}
		\pgfpathlineto{\pgfpointadd{\northeast}{\pgfpoint{-0.5\pgf@xa}{-0.6\pgf@ya}}}
		\pgfpathlineto{\pgfpointadd{\northeast}{\pgfpoint{-0.33\pgf@xa}{-0.6\pgf@ya}}}
	}
}
\makeatother
\tikzstyle{saturation} = [draw, record, fill=white, minimum height=3em, minimum width=3em]

\tikzstyle{spring}=[thick,decorate,decoration={zigzag,amplitude=0.2cm,pre length=0.3cm,post length=0.3cm,segment length=6}]
\tikzstyle{damper}=[thick,decoration={markings,  
  mark connection node=dmp,
  mark=at position 0.5 with 
  {
    \node (dmp) [thick,inner sep=0pt,transform shape,rotate=-90,minimum width=15pt,minimum height=3pt,draw=none] {};
    \draw [thick] ($(dmp.north east)+(2pt,0)$) -- (dmp.south east) -- (dmp.south west) -- ($(dmp.north west)+(2pt,0)$);
    \draw [thick] ($(dmp.north)+(0,-5pt)$) -- ($(dmp.north)+(0,5pt)$);
  }
}, decorate]
\tikzstyle{ground}=[fill,pattern=north east lines,draw=none,minimum width=0.75cm,minimum height=0.3cm]

\tikzset{
	pics/.cd,
	vector out/.style={
		code={
			\draw[#1] (0,0)  circle (1) (45:1) -- (225:1) (135:1) -- (315:1);
		}
	}
}
\tikzset{
	pics/.cd,
	vector in/.style={
		code={
			\draw[#1] (0,0)  circle (1);
			\fill[#1] (0,0)  circle (.1);
		}
	}
}

\tikzstyle{circlenode} = [draw, thick, circle, minimum width=10pt]

\renewcommand{\baselinestretch}{0.95}

\usepackage[compact]{titlesec}

\titlespacing{\section}{0pt}{1.0ex plus .5ex minus .2ex}{0.7ex plus .5ex minus .2ex}
\titlespacing{\subsection}{0pt}{1.0ex plus .5ex minus .2ex}{0.7ex plus .5ex minus .2ex}
\titlespacing{\subsubsection}{0pt}{1.0ex plus .5ex minus .2ex}{0.7ex plus .5ex minus .2ex}

\makeatletter
\let\NAT@parse\undefined
\makeatother
\usepackage{hyperref}
\ifpdf
  \usepackage{hyperref}
\else
  \usepackage[hypertex]{hyperref}
\fi

\definecolor{dgreen}{rgb}{0,0.7,0}
\definecolor{dblue}{rgb}{0,0,0.60}
\definecolor{dred}{rgb}{0.7,0,0}
\hypersetup{bookmarksnumbered=true,
            plainpages=false,
            colorlinks=true,
            linkcolor=dblue,
            citecolor=dblue,
            filecolor=dblue,
            urlcolor=dblue,
            pdfsubject={Limited circulation access, author: Jon Škerlj, 07. 09. 2026}
}

\algnewcommand\algorithmicswitch{\textbf{switch}}
\algnewcommand\algorithmiccase{\textbf{case}}
\algdef{SE}[SWITCH]{Switch}{EndSwitch}[1]{\algorithmicswitch\ #1}{\algorithmicend\ \algorithmicswitch}%
\algdef{SE}[CASE]{Case}{EndCase}[1]{\algorithmiccase\ #1:}{}%
\algtext*{EndSwitch}%
\algtext*{EndCase}%

\title{\LARGE \bf
Sub-Millisecond Reflexes for Continuous Physical Human-Robot Interaction
}
\author{Jon \v{S}kerlj\textsuperscript{1$\dagger$} and Sami Haddadin\textsuperscript{2}
\thanks{
This work was supported by the Lighthouse Initiative Geriatronics by StMWi Bayern (Project X, grant no. IUK-1807-0007// IUK582/001) and LongLeif GaPa gGmbH (Project Y).}
\thanks{$^{1}$Munich Institute of Robotics and Machine Intelligence (MIRMI), Technische Universität München (TUM), Germany.}
\thanks{$^{2}$Mohamed Bin Zayed University of Artificial Intelligence, UAE}
\thanks{
\textsuperscript{$\dagger$}Corresponding author. Email: \texttt{jon.skerlj@tum.de}}
}
\begin{document}
\maketitle
\thispagestyle{empty}
\pagestyle{empty}


\begin{abstract}
Continuous physical human-robot interaction requires robots to rapidly adapt their control behaviour without interrupting task execution. However, existing ROS 2 control architectures rely on stop-to-switch mechanisms that introduce substantial delays and interrupt execution. We present an open-source, sub-millisecond \href{https://github.com/RoboRef/franka_reflex_engine}{reflex engine} based on an enhanced ROS 2-compatible multimode controller that enables deterministic switching between control modalities and integrates reflex behaviours for mitigating contact hazards and kinematic limit violations. Experiments on a Franka Robotics manipulator demonstrate switching delays below $0.1\:\text{ms}$, more than four orders of magnitude faster than conventional \texttt{ros2\_control}, while maintaining continuous operation during safety-critical events. In a multi-stage manipulation task, the proposed framework reduced execution time by $3.5\%$ and peak trajectory jerk by more than threefold. By eliminating runtime middleware overhead, controller transitions are implemented as lock-free pointer operations, enabling reflex-level responsiveness within an open-source robotics framework.
\end{abstract}

\section{Introduction}
The deployment of service robots in domestic and home environments creates a need for physical human-robot interaction (pHRI) that is inherently safe and robust \cite{gonzalez2011service, skerlj2024}. Unlike a structured lab environment or an industrial cell, a home is a dynamic and unpredictable setting. While future humanoid service robots are expected to operate with human-level capabilities to execute complex, contact-rich manipulation tasks \cite{Ott2006Two-armSystem, Gu2026HumanoidsReview}, there is still work to be done, with current state-of-the-art systems primarily serving as research platforms \cite{pages2016tiago, shigemi2018humanoid, Asfour2018, schwarz2021nimbro, trobinger2021introducing}.

During operation in such surroundings around humans, unforeseen collisions become inevitable \cite{Vorndamme2025ReflexEngine}. Actively mitigating the severity of injuries during these collisions requires torque-controlled systems capable of rapid, low-latency reactions \cite{Haddadin2012InjuryKnowledge, Haddadin2016pHRI}. While safety standards specify the high-rate measuring frequencies, such as 1 kHz (ISO 10218 \cite{ISO10218}, ISO 13482 \cite{ISO13482}), for collision detection, a major gap remains in translating advanced safety and interaction algorithms to modular, open-source architectures \cite{Vorndamme2025ReflexEngine}.

In addition, the selected range of diverse domestic tasks, e.g., surface cleaning, folding clothes, or close-proximity physical assistance, directly dictates the requirements of the system's real-time control architecture. Different phases of these tasks often impose conflicting control objectives, such as accurate free-space motion, compliant interaction, force regulation, or safety-oriented recovery behaviours \cite{Barreiros2025SciRob}. While unified control formulations can address several of these objectives at once \cite{AlbuSchaeffer2007Unified, Haddadin2024UFIC}, no single control strategy is universally optimal for all phases of service robotic tasks. Different approaches may prioritize accuracy, compliance, force regulation, or rapid safety reactions, requiring distinct control characteristics and tuning objectives. Hence, enabling service humanoids to execute these multifaceted routines successfully requires the capability to dynamically and instantaneously switch their control behaviour according to task demands and environmental conditions \cite{AlbuSchaeffer2007Unified, Barreiros2025SciRob, skerlj2026}.

Nowadays, the prevalent state-of-the-art open-source middleware like Robot Operating System 2 (ROS 2) \cite{Macenski2022ROS2} and its hardware-agnostic control framework \texttt{ros2\_control} are constrained by their reliance on lifecycle node state transitions with a conservative ``stop-to-switch'' implementation \cite{ros_control}. This introduces timing inconsistencies and additional communication overhead \cite{Gambo2025}, which fails to fulfill the real-time requirement and introduces time delays for long-horizon tasks. Consequently, conventional safety mechanisms may completely halt the robot's motion upon safety-critical events. By doing so, these approaches sacrifice fluid physical manipulation and system autonomy for functional safety \cite{Vorndamme2025ReflexEngine}. 

\begin{figure}[t]%
	\centering
    \includegraphics[width=0.99\linewidth]{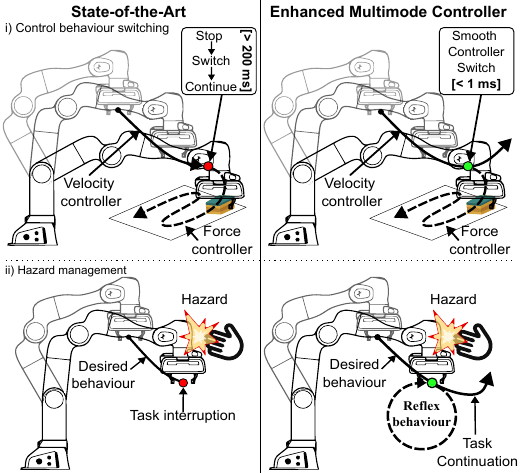}
	\caption{Conventional control frameworks interrupt tasks during control mode switching and hazardous events (left), whereas the proposed framework enables sub-millisecond reflex responses and uninterrupted physical human-robot interaction (right).}
	\label{fig:intro}
\end{figure}

Having established the need for multi-modal control, safety requirements of low-latency interactions, and the limitations of the middleware ``stop-to-switch" bottleneck, as visualized in Fig.~\ref{fig:intro}, the novel contributions of our work are:
\begin{itemize}
    \item A \textbf{real-time, sub-millisecond} controller switching architecture for smooth continuous task execution compatible with existing ROS 2 ecosystems.
    \item A \textbf{reflex engine} built upon the proposed switching architecture, enabling rapid reactions to contact hazards, joint limits, and excessive velocities before manufacturer safety stops are triggered.
    \item \textbf{Continuous execution} of multistage manipulation tasks through seamless transitions between control behaviours, eliminating the traditional stop-to-switch interruption.
    \item \textbf{Empirical validation} on a physical Franka Robotics manipulator, demonstrating switching latencies below $0.1\:\text{ms}$, successful hazard mitigation, and uninterrupted task execution.
\end{itemize}

\section{State of the Art}
\subsection{High-Frequency Torque Control}\label{sec:sota_torque_control}
The groundwork for robots to be able to work side by side with humans today was established with the introduction of force and torque control \cite{Whitney1977}. The early work in this area supported control frequencies of 60 Hz \cite{Salisbury1980} and 120 Hz \cite{Craig1981}, which prevented them from detecting higher oscillations at around 280 Hz. With the rapid technological advancement in recent decades, the current safety standards, e.g., ISO 10218 \cite{ISO10218}, require 1 kHz control and measurement frequencies to detect collisions and minimize injury risks during pHRI tasks. This torque command loop frequency enables utilization of Lagrangian dynamics for modelling, so that the joints can be treated as ideal torque sources by higher-level controllers  \cite{Haddadin2022RefPlat}. Consequently, we have seen a prevalent increase in available collaborative robots (cobots), such as the DLR Lightweight Robot Series \cite{Bischoff2010}, Franka Emika Panda \cite{Haddadin2022RefPlat} \& FR3 \cite{FR32026}, KINOVA Gen 3 \cite{Kinova2022}, etc.

\subsection{Reflex Behaviours}\label{sec:sota_reflex}
A continuous, deterministic $1$ kHz torque-control loop provides the necessary sensory readings for collision detection by estimating external contact wrenches, $\mathcal{F}_{\text{ext}}$, via proprioceptive observers \cite{DeLuca2006CollisionDetection, Haddadin2008CollisionDetection}. However, translating these torque observations into safe physical interaction requires a structured control response tailored to the physical evolution of a collision \cite{Vorndamme2025ReflexEngine}. In the literature, physical human-robot impacts are categorized into four distinct temporal phases \cite{Vorndamme2025ReflexEngine}: 
\begin{enumerate}
\renewcommand{\theenumi}{\roman{enumi}}
    \item\textbf{Pre-Collision Phase:} Anticipatory measures (velocity scaling, posture optimization) bound kinetic energy before contact \cite{Haddadin2012InjuryKnowledge}.
    \item\textbf{Impact Phase I (Transient):} Open-loop dynamics ($10$--$20\:\text{ms}$) governed by structural inertia. Feedback control cannot attenuate initial peak forces \cite{Haddadin2007, Haddadin2008CollisionDetection}.\item\textbf{Impact Phase II (Quasi-Static):} Slower contact dynamics where low-latency control eliminates residual crushing forces and clamping \cite{Haddadin2010Soft-tissue, Haddadin2012InjuryKnowledge}.
    \item\textbf{Post-Collision Phase:} Reaction termination, secondary hazard management, and task recovery \cite{Parusel2011State-based_behaviour, Vorndamme2025ReflexEngine}.
\end{enumerate}
Since feedback loops cannot influence Phase~I of the collision, reflex behaviours aim to mitigate the consequences of the last two stages. These maneuvers are categorized by their degree of task modification \cite{Vorndamme2025ReflexEngine}:
\begin{enumerate}
\renewcommand{\theenumi}{\roman{enumi}}
    \item \textbf{Task-Preserving Reflexes:} e.g., by adjusting execution timing ($\Delta t$) along the nominal path without altering spatial geometry \cite{Haddadin2008CollisionDetection}.
    \item \textbf{Task-Relaxing Reflexes:} e.g., by permitting transient path shifts before resuming normal tracking \cite{Vorndamme2025ReflexEngine}.
    \item \textbf{Task-Abandoning Reflexes:} e.g., by suspending tracking upon reaching the safety limits \cite{Vorndamme2025ReflexEngine}.
\end{enumerate}
Despite the maturity of collision detection, temporal impact phases, and reflex taxonomies, their implementation has historically been confined to custom, proprietary real-time frameworks, for instance DLR's aRD/Beasty \cite{Ott2006Two-armSystem, Parusel2011State-based_behaviour} or ArmarX-RT \cite{Asfour2018}, which lack generalizability and open-source accessibility. To achieve broader adoption across the robotics community, compatibility with modern open-source middlewares is highly desirable.

Beyond reactive control, monitoring and recording frameworks have been proposed to capture robot and human states during pHRI, allowing safety-critical interactions to be analyzed and reconstructed \cite{skerlj2023VR, skerlj2024DataRec}. Such approaches provide complementary observability and traceability, whereas the present work focuses on closing the safety loop online through low-latency detection, reflex activation, and recovery.
\subsection{Existing Robotics Middlewares}\label{sec:sota_middleware}
Modular robotics middlewares, primarily ROS 2 \cite{Macenski2022ROS2} and its hardware-agnostic control framework \texttt{ros2\_control} \cite{ros_control}, have greatly simplified robotics development utilizing component-based architectures. However, this modularity introduces significant communication and scheduling bottlenecks. Standard distributed multi-process configurations suffer from quadratic memory growth ($\mathcal{O}(N^2)$) due to DDS network participant discovery \cite{Macenski2023}; meanwhile, serialization overhead and default executor scheduling models introduce timing jitter and latency \cite{Macenski2022ROS2, Macenski2023}. To address these problems, ROS 2 introduced node composition and intra-process communication, which uses smart pointers to achieve stable microsecond-level latencies ($\approx 40\;\mu\text{s}$) \cite{Macenski2023}. 

Despite these optimizations, the standard \texttt{ros2\_control} architecture remains heavily constrained by relying on standard lifecycle node transitions and a conservative ``stop-to-switch" paradigm  that involves various service calls and resource allocations \cite{ros_control} that block the real-time loop. To minimize velocity discontinuities during controller switching, some approaches use predefined transition zones to blend from one controller to another \cite{Mejia2025}. Unfortunately, this approach could lead to dangerous outcomes in contact-rich pHRI tasks where potential transition zones are almost non-existent. Conversely, the \texttt{multipanda\_ros2} addressed this issue to a certain degree, taking advantage of referencing and shared memory \cite{skerlj2026}. However, the reported values still showed an average switching delay of approximately $2\: \text{ms}$, which still does not satisfy strict ISO-defined requirements for collision detection \cite{ISO10218, ISO13482, ISO15066}.

On the other hand, high-performance platforms, to this day, usually rely on proprietary or in-house built frameworks to achieve deterministic execution. For instance, the DLR humanoid Rollin' Justin uses custom-built controllers for real-time control \cite{Ott2006Two-armSystem}. Similarly, the OROCOS framework integrates specialized toolkits to guarantee deterministic execution \cite{Bruyninckx2001Orocos, Elkady2012Middleware}. While these custom solutions meet the strict timing constraints, they lack plug-and-play modularity, adaptability, and generalizability found in modern open-source ecosystems. This forces an unwanted trade-off between the flexibility of general-purpose middlewares and the rigid determinism of platform-specific architectures \cite{Elkady2012Middleware, Macenski2022ROS2}.

\section{Methods}\label{sec:methods}
\subsection{Multimode Controller Architecture}
The foundation of the proposed real-time behaviour switching is built upon the open-source $multipanda\_ros2$ framework, which provides native ROS 2 interfaces for coordinated control of multiple robots from a single process\cite{skerlj2026}. At its core is the Multimode Controller (MC), which integrates seamlessly into the ROS 2 environment as a standard \texttt{ros2\_control} controller. 

Rather than executing a monolithic control law, the MC acts as a central supervisor that orchestrates a modular suite of lightweight subprocesses called \textit{controllets}. The MC itself does not compute a control law; instead, it delegates this task to the active controllets. Each controllet is assigned ownership of one or more hardware resources in the \texttt{ros2\_control} framework at runtime and implements the specific command torque $\boldsymbol{\tau}_{\text{cmd}}$ calculation required for its assigned task. The MC can be instantiated with an arbitrary number of these controllets based on the user's configuration, strictly ensuring that no more than one controllet has ownership of any single physical motor at any given instant to prevent hardware resource conflicts.

To achieve deterministic execution and minimize computational overhead within the real-time loop, the architecture is designed to completely avoid runtime memory allocations and unnecessary data copying. During initialization, the MC is populated with a vector of configured controllets. Each controllet is constructed utilizing C++ move operators and holds direct references to the shared command vector $\boldsymbol{\tau}_{\text{cmd}}$ and the robot state data (e.g., joint positions and velocities) centrally managed by the MC. This zero-copy reference design provides controllets with instantaneous, real-time access to the complete robot state, which guarantees that the calculated torque commands remain temporally consistent when written to the underlying low-level robot control loop. For further details, please refer to \cite{skerlj2026}.

\begin{figure}[t]%
	\centering
    \includegraphics[width=0.90\linewidth]{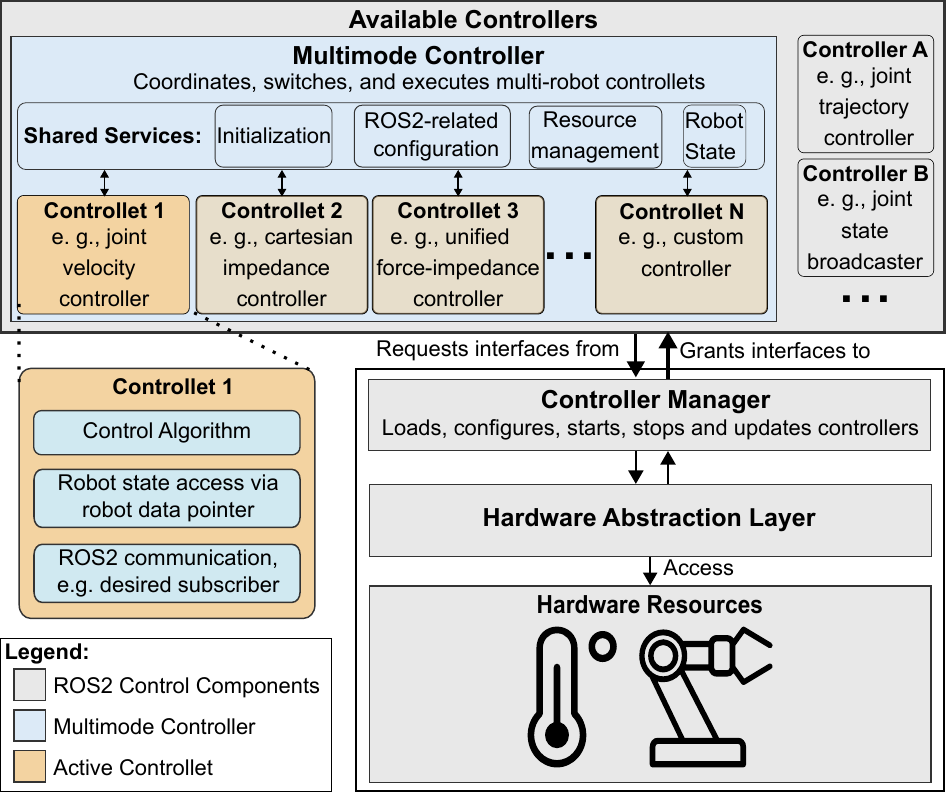}
	\caption{Visualization of the improved multimodal controller within the \texttt{ros2\_control} framework. More detail about \texttt{ros2\_control} components can be found in \cite{ros_control}.}
	\label{fig:mc_architecture}
\end{figure}

\subsection{Optimization for Sub-Millisecond Behaviour Switching}
Maintaining a strict 1~kHz control frequency and deterministic sampling rate is an essential requirement for safe pHRI \cite{ISO10218, Vorndamme2025ReflexEngine}. The original implementation of the framework, transitioning between active controllets, incurs a switching delay of approximately $2\text{ ms}$ \cite{skerlj2026}. 

Detailed profiling and system-level tracing revealed that this multi-millisecond delay was primarily caused by the dynamic creation of ROS 2 communication entities during runtime. For instance, when a ROS 2 subscriber is instantiated, the corresponding callback registration and middleware resources must also be initialized, which is highly non-deterministic because it relies on the underlying Data Distribution Service (DDS) \cite{Macenski2022ROS2}. It may trigger dynamic heap memory allocations, potentially leading to physical memory page faults, requiring internal synchronization locks within the middleware abstraction layers to update the global node graph and execute peer-to-peer participant discovery protocols \cite{Macenski2022ROS2, Gambo2025}. When these operations occur within or interfere with high-priority real-time execution threads, they can produce substantial execution jitter, thread blocking, and control loop deadline misses, thereby violating hard real-time requirements.

To achieve sub-millisecond determinism and enable instantaneous reflex reactions within open-source ecosystems, all ROS 2-related instantiations and communication interface bottlenecks were relocated to the timing-insensitive initialization phase of the MC. By pre-allocating memory and establishing all middleware endpoints before the real-time execution loop begins, the system completely avoids runtime heap allocations and expensive node graph modifications on the fly. Within the 1~kHz real-time execution thread, switching the active controllet is now reduced to a deterministic, lock-free pointer reassignment within the active memory space. Since both active and standby controllets maintain pre-initialized, direct references to the centrally managed robot state data and the shared command vector $\boldsymbol{\tau}_{\text{cmd}}$, changing control modalities does not require any context switches, system calls, or middleware intervention. This zero-overhead pointer swap successfully preserves the execution of the 1~kHz control cycle during transitions, eliminating middleware-induced delays.

\subsection{Reflex Behaviours}
\label{sec:preemptive_reflexes}
While standard collaborative robot controllers rely on manufacturer-level safety monitors \cite{Haddadin2022RefPlat} that lock brakes upon limit violations, these rigid emergency stops cause severe hardware deadlocks, disrupt real-time control, and risk human injury from unmitigated clamping forces during post-impact phases. To ensure continuous operational safety and prevent rigid hardware stops, our framework embeds $1\,\text{kHz}$ \emph{reactive reflex behaviours} into the multimode controller framework, capitalizing on its features. By combining instantaneous torque damping with $< 1\,\text{ms}$ asynchronous controllet switching, the engine intercepts external contact forces, joint kinematic limits, and excessive velocities before manufacturer thresholds are reached, providing compliant collision mitigation, object affordance preservation, and seamless haptic task recovery\footnote{The source code with examples and documentation is publicly available at: \url{https://github.com/RoboRef/franka_reflex_engine}.}.

\subsubsection{Hazard Detection and Interception}
At each $1\,\text{kHz}$ cycle, joint-level position ($q$), velocity ($\dot{q}$), and torque ($\tau$) safety boundaries are evaluated for each joint $i \in \{1, \dots, n\}$, where $n$ denotes the total number of degrees of freedom:
\begin{itemize}
    \item \textbf{Position Limits ($\mathcal{H}_{\text{pos}}$):} With buffer vector $\boldsymbol{\delta}_q  >0$:
    \begin{equation}
        \mathcal{H}_{\text{pos}_i} = 
        \begin{cases} 
        1, & \text{if } q_i \notin [q_{\min, i} + \delta_{q_i}, \; q_{\max, i} - \delta_{q_i}], \\
        0, & \text{otherwise}.
        \end{cases}
    \end{equation}
    \item \textbf{Velocity Limits ($\mathcal{H}_{\text{vel}}$):} With buffer vector $\boldsymbol{\delta}_{\dot{q}} >0$:
    \begin{equation}
        \mathcal{H}_{\text{vel}, i} = 
        \begin{cases} 
        1, & \text{if } \dot{q}_i \notin [-\dot{q}_{\max, i} + \delta_{\dot{q}_i}, \; \dot{q}_{\max, i} - \delta_{\dot{q}_i}], \\
        0, & \text{otherwise}.
        \end{cases}
    \end{equation}
    \item \textbf{Joint Torque Limits ($\mathcal{H}_{\tau}$):} With buffer vector $\boldsymbol{\delta}_{\tau} >0$:
    \begin{equation}
        \mathcal{H}_{\tau_i} = 
        \begin{cases} 
        1, & \text{if } \hat{\boldsymbol{\tau}}_{\text{ext}} \notin [\boldsymbol{\tau}_{\min_i}, \boldsymbol{\tau}_{\max_i}], \\
        0, & \text{otherwise}.
        \end{cases}
    \end{equation}
\end{itemize}
On the operational space level, at the point of interest (typically the end-effector), the safety boundary is defined by the estimated contact wrench $\mathcal{F}_{\text{ext}} = [\boldsymbol{f}_{\text{ext}}, \boldsymbol{m}_{\text{ext}}]^T = \boldsymbol{J}^{\dagger}\hat{\boldsymbol{\tau}}_{\text{ext}} \in \mathbb{R}^6$:
\begin{itemize}
    \item \textbf{Absolute Force Magnitude ($\mathcal{H}_{f}$):} With scalar buffer $f_{max} > 0$:
    \begin{equation}
        \mathcal{H}_{f} = 
        \begin{cases} 
        1, & \text{if } \|\boldsymbol{f}_{\text{ext}}\| > f_{\max}, \\
        0, & \text{otherwise}.
        \end{cases}
    \end{equation}
    \item \textbf{Absolute Moment Magnitude ($\mathcal{H}_{m}$):} With scalar buffer $m_{max} > 0$:
    \begin{equation}
        \mathcal{H}_{m} = 
        \begin{cases} 
        1, & \text{if } \|\boldsymbol{m}_{\text{ext}}\| > m_{\max}, \\
        0, & \text{otherwise}.
        \end{cases}
    \end{equation}
\end{itemize}
\subsubsection{Reflex State Machine}

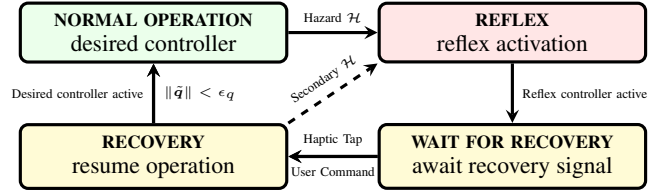
\begin{figure}[t]
\centering
\resizebox{\columnwidth}{!}{%
\begin{tikzpicture}[->, >=stealth, node distance=0.8cm and 1.2cm, auto, very thick,
    base/.style={rectangle, rounded corners=3pt, inner sep=4pt, text width=3.3cm, text centered, font=\scriptsize},
    normal_style/.style={base, draw=black!70!black, fill=green!8},
    hazard_style/.style={base, draw=black!70!black, fill=red!10},
    rec_style/.style={base, draw=black!80!black, fill=yellow!20}]
    
    \node[normal_style] (normal) {\textbf{NORMAL OPERATION}\\ \vspace{2pt}\small desired controller};
    \node[hazard_style, right=of normal] (reflex) {\textbf{REFLEX}\\ \vspace{2pt}\small reflex activation};
    \node[rec_style, below=of normal] (rec) {\textbf{RECOVERY}\\ \vspace{2pt}\small resume operation};
    \node[rec_style, below=of reflex] (wait) {\textbf{WAIT FOR RECOVERY}\\ \vspace{2pt}\small await recovery signal};
    
    \path[->, font=\tiny] 
        (normal) edge node[above] {Hazard $\mathcal{H}$} (reflex)
        (reflex) edge node[right] {Reflex controller active} (wait)
        (wait) edge node[above] {Haptic Tap} (rec)
        (wait) edge node[below] {User Command} (rec)
        (rec) edge node[right] {$\|\tilde{\boldsymbol{q}}\| < \epsilon_q$} (normal)
        (rec) edge node[left] {Desired controller active} (normal)
        (rec.north east) edge[dashed] node[above, sloped] {Secondary $\mathcal{H}$} (reflex.south west);
\end{tikzpicture}%
}
\caption{State machine for reactive reflex behaviours.}
\label{fig:reflex_state_machine}
\end{figure}

The state machine transitions across four discrete states $\mathcal{S} \in \{\textsf{\scriptsize NORMAL}, \textsf{\scriptsize REFLEX}, \textsf{\scriptsize WAIT\_FOR\_RECOVERY}, \textsf{\scriptsize RECOVERY}\}$, evaluated inside the $1\,\text{kHz}$ execution loop visualized in Fig.~\ref{fig:reflex_state_machine}. 

All safety parameters, threshold modes, buffer margins, and reflex behaviours are defined declaratively via ROS~2 YAML configuration files, allowing per-application tuning without recompilation. For an intuitive user experience during program execution, a haptic gesture ($\|\boldsymbol{f}_{\text{ext}}\| > 10\,\text{N}$) feature was implemented to trigger the recovery procedure without the need to run any manual computer commands. After the triggered hazard has been cleared, the desired operation can continue as before.

\subsubsection{Available Reflex Behaviours}
Following the taxonomy introduced in Sec.~\ref{sec:sota_reflex}, our reflex engine integrates the following five task-abandoning reflex behaviours (Tab.~\ref{tab:reflex_control_laws}):
\begin{enumerate}
\renewcommand{\theenumi}{\roman{enumi}}
    \item \textbf{Stop Reflex:} Freezes the manipulator at the time of hazard detection $\boldsymbol{q}(t_{\text{cd}})$.
    \item \textbf{Admittance Reflex:} Maps measured external torques $\hat{\boldsymbol{\tau}}_{\text{ext}}$ to reference joint velocities via a proportional gain $\alpha$, guiding compliant evasive motion along the direction of external interaction forces.
    \item \textbf{Joint Retraction:} Displaces the target joint configuration $\boldsymbol{q}_d^*$ along the negative joint torque vector $-\hat{\boldsymbol{\tau}}_{\text{ext}}$ scaled by Cartesian bounds $\delta = 0.05\,\text{m}$ via the manipulator Jacobian, actively relieving joint-level contact stresses.
    \item \textbf{Cartesian Retraction:} Actively displaces the Cartesian target position $\boldsymbol{x}_d^*$ by distance $\gamma$ along the negative external contact force vector, $-\boldsymbol{f}_{\text{ext}}$, while maintaining fixed orientation to prevent object tilting and secondary hazards.
    \item \textbf{Zero-G Reflex:} Commands pure gravity compensation, eliminating residual contact forces and preventing severe human clamping in constrained collision scenarios.
\end{enumerate}

\begin{table}[h]
\centering
\caption{Available state-of-the-art reflex behaviours~\cite{Vorndamme2025ReflexEngine}.}
\label{tab:reflex_control_laws}
\resizebox{\columnwidth}{!}{%
\renewcommand{\arraystretch}{1.5}
\begin{tabular}{@{}lll@{}}
\hline\hline
\textbf{Name} & \textbf{Control Law} & \textbf{Parameters} \\ \hline
\textbf{Stop} & $\boldsymbol{q}_d^*(t) = \boldsymbol{q}(t_{\text{cd}})$ & \\
\textbf{Admittance} & $\dot{\boldsymbol{q}}_d^*(t) = \alpha \hat{\boldsymbol{\tau}}_{\text{ext}}(t)$ & $\alpha = 0.05$ \\
\textbf{Joint Retract} & $\boldsymbol{q}_d^*(t) = \boldsymbol{q}(t_{\text{cd}}) - \delta \frac{\hat{\boldsymbol{\tau}}_{\text{ext}}(t_\text{cd})}{\|\boldsymbol{J}_{cart}\hat{\boldsymbol{\tau}}_{\text{ext}}(t_\text{cd})\|}$ & $\delta = 0.05$ \\
\textbf{Cart. Retract} & $\boldsymbol{x}_d^*(t) = \boldsymbol{x}(t_{\text{cd}}) - \begin{bmatrix} \gamma \frac{\boldsymbol{f}_{\text{ext}}}{\|\boldsymbol{f}_{\text{ext}}\|} \\ \mathbf{0} \end{bmatrix}$ & $\gamma = 0.09$ \\
\textbf{Zero-G} & $\boldsymbol{\tau}_d(t) = \boldsymbol{\tau}_g(\boldsymbol{q})$ &  \\
\hline\hline
\end{tabular}%
}
\end{table}
\section{Experimental Procedure}\label{sec:experiments}
All experiments were conducted on a Franka Robotics Panda manipulator controlled by an ASRock Industrial 4X4 Box 7735U/D5 NUC mini-PC with $32\:\text{GB}$ of DDR5-4800 MHz RAM. The underlying software architecture was built on ROS 2, utilizing the open-source \texttt{multipanda\_ros2} framework \cite{skerlj2026} as the real-time backbone. To validate the proposed architecture, the experimental procedure was divided into three phases: i) Quantify the control switching delay, ii) Evaluate the reactive reflex behaviours, and iii) Execute a multi-stage task to show the capabilities compared to the state-of-the-art.

\subsection{Control Behaviour Switching Delay}\label{sec:exp_control_beh}
To benchmark the latency of state transitions, we measured the switching delay across 5 control behaviours mentioned in Tab.~\ref{tab:reflex_control_laws} for three distinct configurations:
\begin{enumerate}
\renewcommand{\theenumi}{\roman{enumi}}
    \item Baseline \texttt{ros2\_control}: Standard lifecycle state transitions, switching an active controller to an inactive state and vice versa with all controllers already loaded and configured at the start of the experiment.
    \item Original multimode controller: The baseline  \texttt{multipanda\_ros2} framework using the default controllet switching.
    \item Improved multimode controller with optimized software design to minimize external delays at runtime.
\end{enumerate}
Each configuration was recorded for a total of $100$ switches of the control behaviour. The switching delay was quantified by recording the exact timestamp when the switch initiation request was triggered and comparing it against the timestamp when the newly computed command values, $\tau_{cmd}$, first appeared in the internal libfranka control loop. Cross-correlation was applied between the two recorded datasets to precisely identify the temporal delay for each configuration.

\subsection{Reflex Behaviours}\label{sec:exp_reflex}
To demonstrate the system's ability to intercept hardware faults and maintain continuous operation, a series of kinematic and contact limit tests were conducted.
\begin{enumerate}
\renewcommand{\theenumi}{\roman{enumi}}
    \item \textbf{Joint Position Limit Test}: A stiff joint impedance controller ($K = 800\; \text{N/m}$) was utilized to purposefully command the manipulator into its physical joint limits, with only one joint being actively commanded per experiment. Maximal commanded speed was set at $75\%$ of the hardware limit. This procedure was executed 10 times for each joint across all 7 degrees of freedom for a total of 70 repetitions. The position safety limit was set to $0.2\:\text{rad}$ away from the hardware limits for all joints ($\boldsymbol{\delta}_{q} = 0.2\:\text{rad}$). 
    \item \textbf{Joint Velocity Limit Test}: Mirroring the previous setup, this test utilized a joint velocity controller to aggressively accelerate the manipulator beyond the manufacturer's specifications, maintaining the individual joint per experiment methodology. The velocity safety limit was set to $0.1\;\text{rad/s}$ away from the hardware limits for all joints ($\boldsymbol{\delta}_{\dot{q}} = 0.1\:\text{rad/s}$).
    \item \textbf{Contact Detection}: A mid-level stiffness  ($K = 400\; \text{N/m}$) Cartesian impedance controller was used to command the end effector into a rigid surface with varying end-effector velocity magnitudes, from $0.09$ to $0.26$ m/s before collision. This was repeated 10 times across three different approach configurations to evaluate the consistency of the external force-based reflex trigger. The external force threshold ($\boldsymbol{f}_{\text{max}}$) was set to $15\:\text{N}$ for all experiments. 
\end{enumerate}
For position and velocity experiments, a stop reflex behaviour was implemented via a soft joint impedance controller (K$_{1-4} = 200\:\text{N/m}$, K$_{5-7} = 100\:\text{N/m}$). However, contact detection tests utilized a Zero-G approach to resolve unexpected collisions.
\subsection{Multistage Task Execution}\label{sec:exp_task}
To evaluate the real-world operational impact of real-time control-behaviour switching, we designed a continuous, multi-phase manipulation task inspired by a real use case where the robot is instructed to clean the table. This scenario was divided into three distinct parts, between which control behaviour must be switched:
\begin{enumerate}
\renewcommand{\theenumi}{\roman{enumi}}
\item Pick \& Place: the robot relocates objects out of the way by moving quickly and precisely, utilizing a fast velocity controller.
\item Surface wiping: requires the robot to maintain a steady normal force of $10$ N against a surface while following a trajectory along a plane via a unified force-motion controller according to \cite{Haddadin2024UFIC}.
\item Tool Release \& Return: the robot returns the sponge to its designated spot and goes back to the starting position using a stiff joint impedance controller ($K = 800\; \text{N/m}$).
\end{enumerate}
The task performance of the enhanced multimode controller was benchmarked against the state-of-the-art \texttt{ros2\_control} baseline. The task was executed 10 times for each approach, with evaluation metrics being \textbf{task completion time} and \textbf{peak trajectory jerk}.

\section{Results}

\subsection{Control Behaviour Switching Delay}\label{sec:res_switching_delay}

\begin{figure}[H]%
	\centering
    \subfloat[\centering Baseline \textit{ros2\_control} \label{fig:ros2_control_delay}]{{\includegraphics[width=0.30\columnwidth]{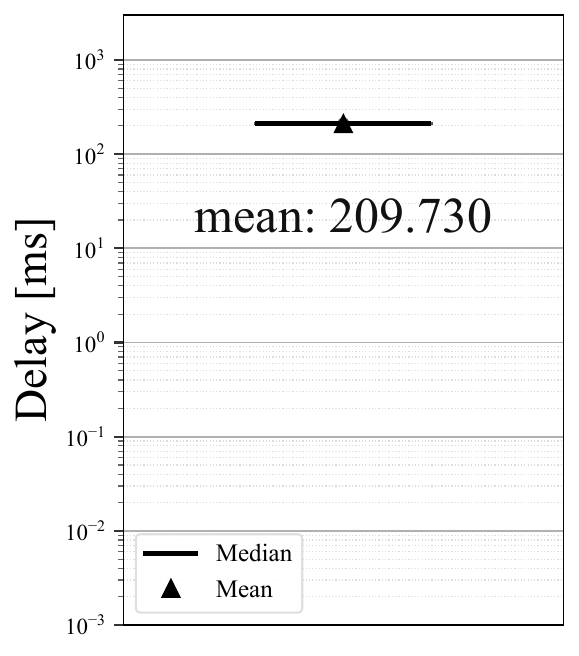} }}%
	\smallskip{}
	\subfloat[\centering Original multimode controller\label{fig:default_MC_delay}]{{\includegraphics[width=0.30\columnwidth]{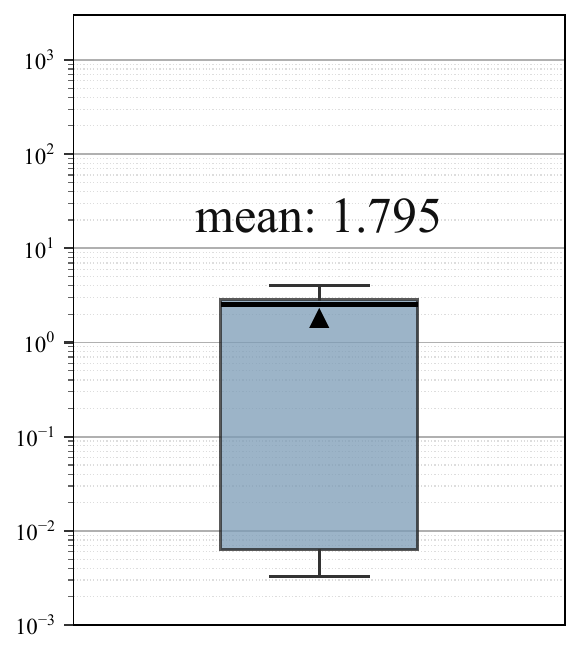} }}
	\smallskip{}
	\subfloat[\centering Improved multimode controller\label{fig:improved_MC_delay}]{{\includegraphics[width=0.30\columnwidth]{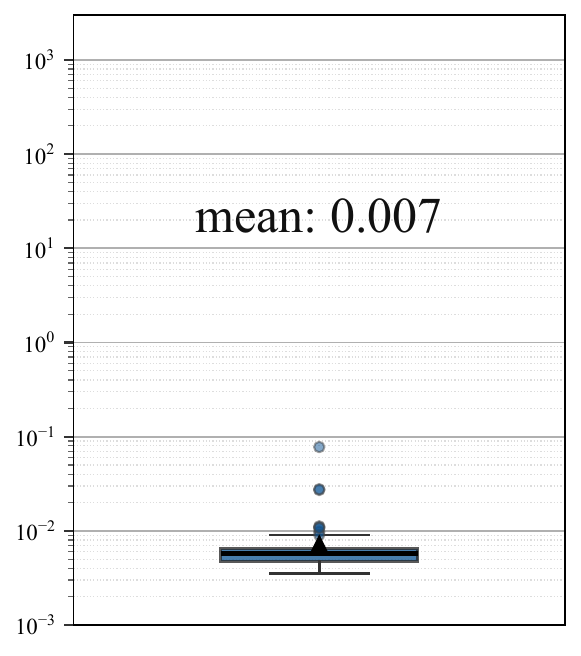} }}%
	\smallskip{}
	\caption{Box plots of controller switching delays using a logarithmic scale for a) baseline \textit{ros2\_control} framework, b) original multimode controller, and c) improved multimode controller for $100$ switches per each configuration.}
	\label{fig:switching_delay}

\end{figure}

The results of the switching latency can be seen in Fig.~\ref{fig:switching_delay}. A logarithmic scale is used to visualize the results, as they span multiple orders of magnitude. On average, the \texttt{ros2\_control} baseline resulted in approximately $210$ milliseconds of delay, while the default multimode controller and its improved version ended with an average delay of $1.8$ and $0.01$ ms, respectively. Furthermore, the latency distribution shown in Fig.~\ref{fig:switching_delay} demonstrates that even the worst observed switching event of our proposed approach remained below $0.1\:\text{ms}$, indicating deterministic behaviour with negligible switching jitter across 100 trials.
\vspace{3mm}

\subsection{Reflex Behaviours}\label{sec:res_reflex}
\begin{figure}[H]%
	\centering
    \subfloat[\centering Joint Position Test \label{fig:pos_limit}]{{\includegraphics[width=0.45\columnwidth]{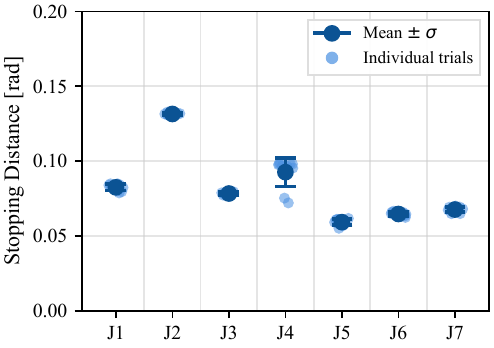} }}%
	\smallskip{}
	\subfloat[\centering Joint Velocity Test \label{fig:vel_limit}]{{\includegraphics[width=0.45\columnwidth]{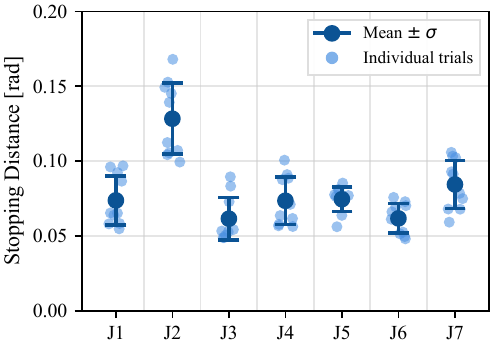} }}
	\caption{Stopping distances displayed per joint for a) joint position test and b) joint velocity test.}
\label{fig:pos_vel_stats}
\end{figure}

The statistics of the stopping distances for the joint position and velocity limit tests are summarized in Fig.~\ref{fig:pos_vel_stats}. The mean and standard deviation of the stopping distance values averaged across all 7 joints are $0.082 \pm 0.003$ and $0.080 \pm 0.016\; \text{rad}$  for the position and velocity limit tests, respectively. It can be seen that the means are comparable while the variability is significantly higher during the joint velocity limit tests due to the higher joint velocities at the time of the reflex trigger. 

\begin{figure}[h]%
	\centering
    \subfloat[\centering Joint Position Test \label{fig:pos_traj}]{{\includegraphics[width=0.75\columnwidth]{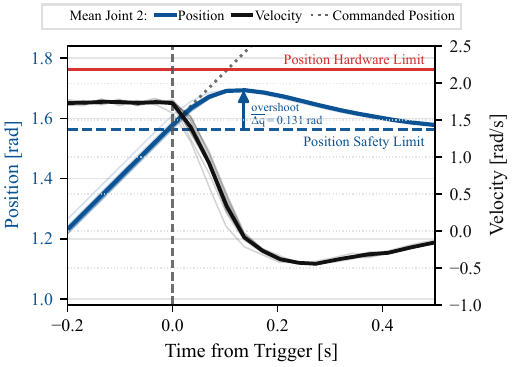} }}%
    \smallskip{}
	\subfloat[\centering Joint Velocity Test \label{fig:vel_traj}]{{\includegraphics[width=0.75\columnwidth]{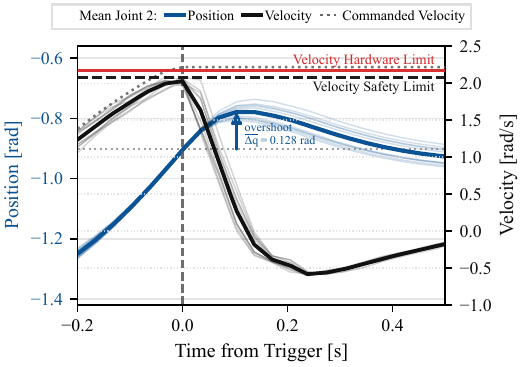} }}
	\caption{Example trajectories of Joint 2 for (a) joint position and (b) velocity limit tests. Recorded joint angles are shown in blue and velocities in black, corresponding to the respective colour-coded vertical axes.}
	\label{fig:pos_vel_traj}
\end{figure}

Moreover, Fig.~\ref{fig:pos_vel_traj} displays the recorded position and velocity trajectories for joint $2$ since it produced the highest stopping distances. In the default mounting, this joint carries most of the robot weight, which creates the most demanding stopping conditions. Even with the highest observed values, none of the trials reached the actual hardware limit, which would completely stop the robot and produce high jerks.

\begin{figure}[H]%
	\centering
    \includegraphics[width=0.5\linewidth]{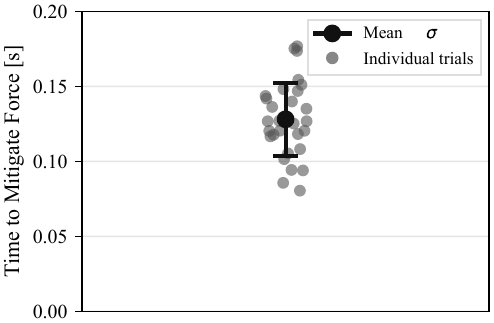}
	\caption{Results of the contact detection tests displaying time to mitigate force after an unexpected collision.}
	\label{fig:force_stats}
\end{figure}

\begin{figure}[H]%
	\centering
    \includegraphics[width=0.75\linewidth]{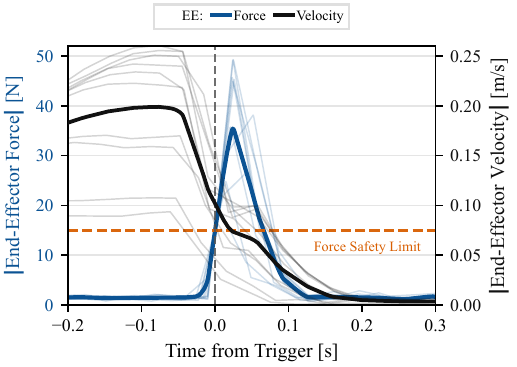}
	\caption{Results of the contact detection tests displaying the magnitudes of end-effector force (blue) and velocity (black). Both trajectories correspond to the respective color-coded vertical axes.}
	\label{fig:force_traj}
\end{figure}

Finally, across $30$ contact detection trials, Fig.~\ref{fig:force_stats} shows a mean force mitigation time of $0.122 \pm 0.013\:\text{s}$ for external forces exceeding $15\:\text{N}$, while Fig.~\ref{fig:force_traj} illustrates the corresponding end-effector stopping behaviours for measured external force (blue) and velocity (black) magnitudes. Similar to before, low-level hardware limits were not triggered, allowing uninterrupted execution throughout the experimental procedure.

\subsection{Multistage Task Execution}\label{sec:res_task}
\begin{figure}[h]%
	\centering
    \includegraphics[width=0.95\linewidth]{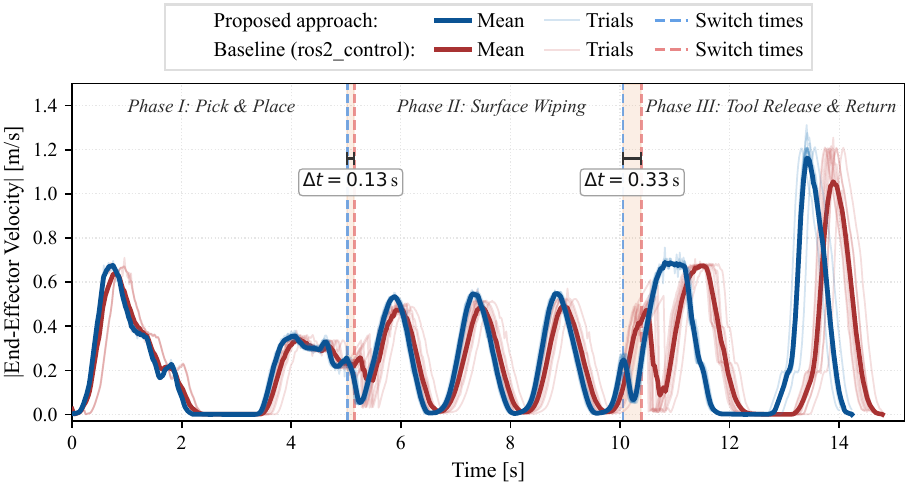}
	\caption{Multi-stage task velocity comparison between state-of-the-art baseline (\textit{ros2\_control}) in red and our approach (enhanced multimode controller) in blue.}
	\label{fig:multistage_task}
\end{figure}

The results of the real-world-inspired task can be seen in Fig.~\ref{fig:multistage_task} and Tab.~\ref{tab:multifaceted_task_comparison}. The task evolution is represented using the magnitude of task space velocity; blue colour denotes our approach, versus the baseline in red. The task completion time metric shows a significant speedup of nearly $0.5\:\text{s}$. Additionally, smoother control behaviour switches utilized by the multimode controller reduced peak operational-space jerk from $499.0 \pm 96.9\:\text{m/s$^3$}$ to $155.4 \pm 40.6\:\text{m/s$^3$}$.

\begin{table}[h]
\centering
\caption{Quantitative performance comparison across 20 trials of the multifaceted task ($N=10$ per configuration).}
\label{tab:multifaceted_task_comparison}
\begin{tabular}{lcc}
\hline
\textbf{Metric} & \textbf{Proposed approach} & \textbf{ros2\_control} \\
\hline
Total Task Duration $[\mathrm{s}]$      & $\mathbf{14.09 \pm 0.06}$ & $14.58 \pm 0.10$ \\
Peak Trajectory Jerk $[\mathrm{m/s^3}]$ & $\mathbf{155.4 \pm 40.6}$ & $499.0 \pm 96.9$ \\
\hline
\end{tabular}
\end{table}
\section{Discussion}
\subsection{Control Behaviour Switching Delay}
\vspace{-2mm}

As explained in Sec.~\ref{sec:res_switching_delay}, the reported switching delay was reduced by more than 4 orders of magnitude compared to the standard open-source \texttt{ros2\_control} baseline. Considering that the improved multimode approach is designed with open-source compatibility in mind, its benefits can be easily utilized by all robotics systems that already use \texttt{ros2\_control} as their main control architecture. The control algorithm must only be copied from a \texttt{ros2\_control} framework as an additional controllet to the multimode controller. In addition, the improved switching latency below $0.1\:\text{ms}$ enables safety-oriented reflex behaviours to execute within the 1 kHz control loop.

\subsection{Reflex Behaviours}
The reflex behaviours of the limit tests to catch manufacturer-defined operation interruptions showed success throughout Sec.~\ref{sec:res_reflex}. Nevertheless, it is important to note that the selection of reflex behaviours and threshold limits ($\boldsymbol{\delta}_{q} = 0.2\:\text{rad}$, $\boldsymbol{\delta}_{\dot{q}} = 0.1\: \text{rad/s}$) plays a major role. For all limit experiments, a stop reflex behaviour was implemented via a soft joint impedance controller (K$_{1-4} = 200\:\text{N/m}$, K$_{5-7} = 100\:\text{N/m}$) that uses the spring behaviour to cushion abrupt stops. Hence, an overshoot behaviour appears that goes beyond the initially determined safety limit. This indicates that depending on the application, careful consideration should be performed when choosing the limits and reflex behaviours.

Lastly, the collision experiments (Fig.~\ref{fig:force_traj}) show that the detected peak transient force ($|\boldsymbol{f}_{\text{max}}|$) is significantly larger than the set collision threshold. In agreement with the physical impact evolution outlined in Sec.~\ref{sec:sota_reflex}, active feedback control cannot attenuate the initial open-loop transient force peak during \textit{Impact Phase I} \cite{Haddadin2008CollisionDetection}. However, because our architecture executes state transitions in sub-millisecond rates, the reflex engages well within \textit{Impact Phase II}, eliminating static crushing forces before lower-level manufacturer safety limiters trip. This prevents sensor saturation and rigid emergency stops historically observed during unexpected rigid impacts \cite{Haddadin2007}.

While the presented experiments focus on repeatable and quantifiable scenarios, supplementary video material demonstrates the reflex engine's reactions during unexpected physical interactions. In these examples, the robot reacts to external contact, transitions to a suitable reflex behaviour, and resumes operation after the hazard has been resolved without triggering low-level safety stops. These examples show the framework's usefulness during realistic pHRI scenarios that are beyond controlled evaluation conditions presented in this work.

\subsection{Multistage Task Execution}
The task described in \ref{sec:exp_task} is a simplified representative real-world scenario to clearly showcase the benefits of the multimode controller. Even with only two control behaviour switches, it already shows significant time savings. The main source of the difference can be seen by looking at Fig.~\ref{fig:multistage_task} in more detail, especially at the second switch time shortly after the $10\:\text{s}$ mark. The task space velocity of our approach is slightly reduced due to the direction change, but it never reaches 0. However, all individual trajectories of the baseline approach are forced to fully stop before the new behaviour can be started. In our use case, this results in a total delay of approx. $0.5\:\text{s}$, which may improve throughput for long-horizon tasks.

Ultimately, the observed jerk in Tab.~\ref{tab:multifaceted_task_comparison} should also be considered before deployment. More than a threefold reduction may reduce mechanical loads during controller transitions.
\section{Conclusion}
This paper presented a novel open-source sub-millisecond controller-switching methodology for safe, uninterrupted task execution without requiring a single unified control strategy. Its modularity and compatibility with the existing ROS 2 ecosystem combine the flexibility of hardware-agnostic frameworks, such as \texttt{ros2\_control}, with timing constraints and determinism typically associated with specialized systems. The proposed approach was validated experimentally on a Franka Robotics manipulator.

The improved multimode controller eliminates runtime overhead associated with modular abstractions and achieves switching delays of $<0.1\:\text{ms}$. This enables control behaviour transitions within latency ranges consistent with controller frequencies commonly required by collaborative robotics safety standards \cite{ISO10218}. Building on this capability, we implemented five reflex behaviours to mitigate critical external forces and intercept joint position and velocity limits before rigid manufacturer safety stops are triggered, thereby maintaining continuous operation. 

Furthermore, a multistage task inspired by a real pHRI use case demonstrated the benefits of seamless switching between control behaviours with different objectives. Despite requiring only two switches, the proposed approach reduced task duration by $3.5\%$ compared with the standard \texttt{ros2\_control} approach. Smoother transitions also reduced the maximum detected trajectory jerk by more than threefold, highlighting benefits for both long-term continuous operation and motion smoothness. 

The presented experiments are limited to a set of predefined reflex behaviours and safety thresholds, whose selection remains application-dependent. Future work will therefore extend the library of available reflex behaviours and refine controller-specific parameters, with particular emphasis on task-preserving strategies that mitigate hazards while minimizing disruption to ongoing manipulation. Beyond classical control, the framework provides a promising foundation for learned robot policies, where rapid transitions to established control strategies may improve robustness in unexpected situations. 

Ultimately, our goal is not only faster controller switching, but enabling robots to react, recover, and continue operating during physical interaction. By bringing reflex-level responsiveness to open-source robotic middleware, we aim to provide the robotics community with a modular and accessible foundation for more capable, resilient, and continuously operating service robots.
{\scriptsize
\setlength{\itemsep}{-1pt}
\bibliographystyle{IEEEtran}
\bibliography{sections/tidy_bibliography}
}

\end{document}